%% file: arxiv.tex
\documentclass[twocolumn]{article}
\usepackage{amsmath,epsfig}
\pdfoutput=1
\usepackage{amsmath}
\usepackage{bm}
\usepackage{url}
\usepackage{svg}
\usepackage{tikz, graphicx}
\usepackage{standalone}
\usepackage{hyperref}
\usepackage[T1]{fontenc}
\usepackage[utf8]{inputenc}
\usepackage{multirow}
\usepackage{color}
\usepackage{float}
\usepackage{caption}
\usepackage{adjustbox}
\usepackage{subcaption}
\usepackage{booktabs}

\newcommand{\nada}[1]{}

\newcommand{\bold}[1]{#1}

\usepackage{environ}
\makeatletter
\newsavebox{\measure@tikzpicture}
\NewEnviron{scaletikzpicturetowidth}[1]{%
  \def\tikz@width{#1}%
  \begin{lrbox}{\measure@tikzpicture}%
  \BODY
  \end{lrbox}%
  \pgfmathparse{#1/\wd\measure@tikzpicture}%
  \BODY
}

\title{Proba-V-ref: Repurposing the Proba-V challenge\\for reference-aware super resolution}

\author{Ngoc Long Nguyen$^1$ \hspace{0.2 cm} Jérémy Anger$^{1,2}$ \hspace{0.2cm} Axel Davy$^1$ \hspace{0.2 cm} Pablo Arias$^1$ \hspace{0.2 cm} Gabriele Facciolo$^1$}
\date{$^1$Université Paris-Saclay, CNRS, ENS Paris-Saclay, Centre Borelli, France
\\ $^2$Kayrros SAS}
\begin{document}
\maketitle
\begin{abstract}

The PROBA-V Super-Resolution challenge distributes real low-resolution image series and corresponding high-resolution targets to advance research on Multi-Image Super Resolution (MISR) for satellite images.
However, in the PROBA-V dataset the low-resolution image corresponding to the high-resolution target is not
identified.
We argue that in doing so, the challenge ranks the proposed methods not only by their MISR performance, but mainly by the heuristics used to guess which image in the series is the most similar to the high-resolution target. We demonstrate this by improving the performance obtained by the two winners of the challenge only by using a different reference image, which we compute following a simple heuristic.
Based on this, we propose PROBA-V-REF a variant of the PROBA-V dataset, in which the reference image in the low-resolution series is provided,
and show that the ranking between the methods changes in this setting.
This is relevant to many practical use cases of MISR where the goal is to super-resolve a specific image of the series, i.e. the reference is known. The proposed PROBA-V-REF should better reflect the performance of the different methods for this reference-aware MISR problem.
\end{abstract}

\input{contents/intro}
\input{contents/related}
\input{contents/methodology}
\input{contents/numerics}
\input{contents/conclusion}

\section*{Acknowledgements}
This work was supported by a grant from Région Île-de-France. It was also partly financed by IDEX Paris-Saclay IDI 2016, ANR-11-IDEX-0003-02, Office  of Naval research grant N00014-17-1-2552, DGA Astrid project  \mbox{\guillemotleft \ filmer la Terre \guillemotright} \ n\textsuperscript{o} ANR-17-ASTR-0013-01, MENRT. This work was  performed using HPC resources 
from GENCI–IDRIS (grant 2020-AD011011801) and  from the “Mésocentre” computing center of CentraleSupélec and ENS 
Paris-Saclay supported by CNRS and Région Île-de-France (http://mesocentre.centralesupelec.fr/).

\bibliographystyle{plain}
\bibliography{arxiv}

\end{document}

%% file: contents/intro.tex
\section{Introduction}

\begin{figure} 
    \centering
    \subfloat{\includegraphics[width=0.97\linewidth]{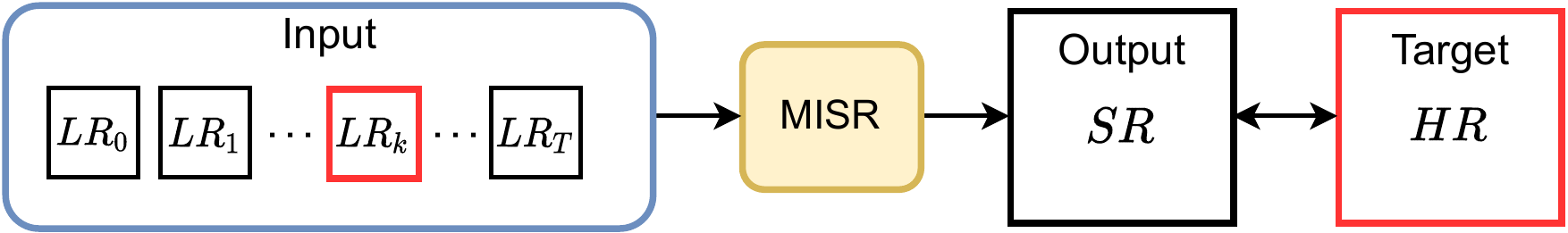} }
    \vspace{3mm}
    \subfloat{\includegraphics[width=0.97\linewidth]{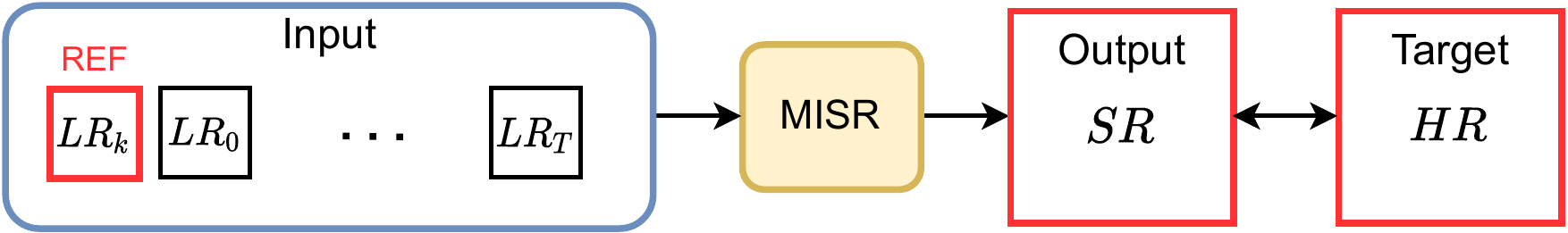} }
    \vspace{2mm}
    \caption{The PROBA-V dataset (top) does not make any distinction between the LR images. One of them was acquired at the same time as the target HR image which is used for training and evaluation.
    The MISR methods need to determine a reference without knowing which is the one corresponding to the target. We propose PROBA-V-REF (bottom), a version of PROBA-V where the identity of true reference is known.}
    \label{fig:1}%
\end{figure}

Earth monitoring plays an important role in our understanding of the Earth systems including climate, natural resources, ecosystems, and natural and human-induced disasters.
Some of Earth monitoring applications require high resolution images, such as monitoring human activity or monitoring deforestation. 
Lately, computational super-resolution is being adopted as a cost-effective solution to increase spatial resolution of satellite images~\cite{murthy2014skysat,anger2020fast}. We refer to~\cite{nasrollahi2014super, yue2016image}
for a comprehensive review of the problem of super-resolution.

In general, the approaches to image super-resolution can be classified into:
single image super-resolution (SISR) and multi-image super-resolution (MISR). 
Single image super-resolution has recently attracted considerable attention in 
the image processing community~\cite{dong2015image, kim2016deeply}.
It is a highly ill-posed problem. In fact, during the acquisition of  the low-resolution (LR) images some high-frequency components are lost or aliased, hindering their correct reconstruction.
In contrast, MISR aims to recover the true details in the super-resolved image (SR) by combining the non-redundant information in multiple LR observations.

In 2019, the Advanced Concepts Team of European Space Agency (ESA) organised a challenge~\cite{martens2019super}
with the goal of super-resolving the multi-temporal images coming from the PROBA-V satellite. 
The challenge dataset consists of sets of LR images acquired within a time window of one month over a set of sites. For each site, a high-resolution target image (HR) is also provided. 
In each sequence, one of the LR images was acquired at the same date as the HR image.
We call this image the true reference.
Knowing the LR reference can help produce a result matching better the HR image as there can be significant changes with images taken at different dates.
However, the identity of these true reference images is not provided in the challenge.
Several teams have participated in the challenge, and since it finished, a ``post-mortem'' contest continues to benchmark new MISR methods. All these works try to solve the problem without the knowledge of the reference images.
We believe that the problem of MISR without a reference image 
is interesting and could have several applications.
However, in such problem, the reference image need to be completely random, which is not the case in the PROBA-V challenge where for example, a cloud-free LR has more chance to be the reference than a cloudy LR image. Such bias introduces noise in the resulting benchmark. A method might get a good performance not because of a more suitable architecture or training, but because of a better heuristic to select the reference image.

On the other hand, reference-aware MISR is a relevant problem in itself. Indeed, in many practical cases, the goal is to super-resolve a specific image of the sequence (for example we might be interested in a specific date). Although this problem is considerably easier, it is far from being solved. 
In other domains such as super-resolution of video or burst of images, the standard definition of the MISR problem includes the reference image.
Hence, we are convinced that a variant of the PROBA-V dataset with the true reference images would be valuable for the computer vision community.

In this work we first demonstrate the impact of the heuristic used to select the reference LR image in the PROBA-V challenge. We do this by improving the performance obtained with the two winning methods of the contest, by simply changing their reference images with a different one chosen following a simple heuristic. We then point out that the true reference image can be obtained in the training and validation splits of the dataset by comparison with the HR target, and propose PROBA-V-REF, a version of the PROBA-V dataset with the true LR references. Finally, we retrain the first and second best methods in the challenge on the proposed PROBA-V-REF dataset and show that the ranking between them becomes inverted.

%% file: contents/related.tex
\section{Related works}
Lately, deep learning algorithms have been proven a success in super-resolution. However, these methods are data-hungry and their performance heavily relies on the quality and the abundance of the training dataset. 

The importance of training  with realistic data was highlighted in~\cite{cai2019toward} for SISR algorithms. The authors of~\cite{cai2019toward} proposed a dataset comprised of real pairs of LR/HR images and showed that the models trained on it  achieved  much better results than those trained on synthetic data~\cite{agustsson2017ntire}.

Realistic MISR datasets are usually small and can only be used to test an MISR algorithm (for example the MDSP dataset~\footnote{http://www.soe.ucsc.edu/\textasciitilde{}milanfar/software/sr-datasets.html}).
Most of deep learning MISR algorithms are trained on simulated data~\cite{wronski2019handheld, masutani2020deep}. 
It was not until the
publication of the PROBA-V dataset that the training of deep learning MISR methods 
could be done on
a real-world dataset.
The PROBA-V satellite is equipped with two types of cameras with different resolutions and revisit times. This interesting setup opens the way to a supervised learning of new MISR methods with real-world data.

However, the limitation of the PROBA-V dataset is that the information of the reference image is not provided, which hinders its huge potential. Indeed, most of traditional MISR methods like shift-and-add, kernel-regression~\cite{takeda2007kernel}, polynomial fitting~\cite{anger2020fast} start by registering all the LR images to a common domain which is usually chosen to be that of one LR image in the series (typically the one we are interested in super-resolving). The two top performing methods of the Proba-V challenge DeepSUM~\cite{molini2019deepsum} and HighRes-net~\cite{deudon2020highres} also pick a specific LR image as an anchor for the reconstruction. DeepSUM selects the LR image with the highest clearance as the reference for the registration step. HighRes-net chooses the median of $9$ clearest LR images as the reference in the fusion step. 

%% file: contents/methodology.tex
\section{Recovering the true LR reference} \label{Sec:Metho}

The PROBA-V dataset contains $566$ scenes from the NIR spectral band and $594$ scenes from the RED band. For each scene, there is only one HR image of $384 \times 384$ pixels and several LR images (from $9$ to $35$ images taken over a period of one month) of $128 \times 128$ pixels. The LR images in one set can be very different due to change of illumination, presence of clouds, shadows or ice/snow covering.
A status map is provided to indicate which pixels in a LR image can be reliable for fusion.
The ``clearance score'' of an image is defined as the percentage of clear pixels in its status map.
The dataset is carefully hand picked such that the LR images have at least
$60\%$ clearance and the HR has at least $75\%$ clearance. Within a 30 day period, even if more than one HR image verify this condition, only the one with the highest clearance is selected as the target. Since the PROBA-V dataset does not make any distinction between the LR images, the MISR methods have to produce some kind of average SR image. To help them recover the true details on the SR image, we need the information of the true LR reference (see Fig.~\ref{fig:1}).

For each element of the training set, we retrieve the true LR reference by determining the LR image that is the most ``similar'' to the HR. To this aim, first a filtered and subsampled (by a factor $3$) version of HR is computed. Then, we align the LR frames with the downsampled HR using the inverse compositional algorithm~\cite{baker2001equivalence}
and compute the pixel-wise root-mean-square errors between them. The true reference is chosen as the LR image that minimizes the error. The computed indexes of the true references for the PROBA-V dataset can be found here: {\small \url{https://github.com/cmla/PROBAVref}}.

%% file: contents/numerics.tex
\section{Experiments}

In this section, we demonstrate that the reference image is as important as the technique used. Then we illustrate and discuss the benefit of the PROBA-V-REF dataset for real-world applications.

For evaluating the quality of the reconstructions we adopt the ``corrected clear'' PSNR (cPSNR)~\cite{martens2019super} metric introduced for PROBA-V challenge. The specificity of this metric is that it takes the status map of the ground truth HR into account and allows intensity biases and small pixel-translations between the super-resolved image and the target.

\subsection{Experimental settings}\label{Sec:quanti}
As mentioned earlier, the two top competitors of the PROBA-V challenge use a specific LR image in the series as anchor.

\noindent \bold{DeepSUM}~\cite{molini2019deepsum} --- is the winner of the challenge. 
It uses the LR with the highest clearance as the reference. A registration step aligns all other images to the reference. 

\noindent \bold{HighRes-net}~\cite{deudon2020highres} --- achieved the second place in the challenge. 
The median of the $9$ images with the highest clearance is considered as a shared representation for multiple LR. Each LR image is embedded jointly with this reference image before being recursively fused.

To show that the choices of the reference images by DeepSUM and HighRes-net are suboptimal we retrain them from scratch using the true LR references (see Sec.~\ref{Sec:Metho}) and name these two adjusted methods \bold{DeepSUM-ref} and \bold{HighRes-net-ref} respectively. Furthermore, we demonstrate that a SISR algorithm trained on the true references can achieve better score than 
DeepSUM and HighRes-net. To this aim, 
we introduce \bold{DeepSUM-SI}, a version of DeepSUM modified to perform SISR by replacing all input images by the true references.

Tables~\ref{tab1} and~\ref{tab2} show the performances of these methods on the validation set for the NIR spectral band,
consisting of 170 scenes.

We consider different ways of choosing the reference on the validation set:

\noindent \textbf{Similarity} --- is the true reference as computed in Sec.~\ref{Sec:Metho}.

\noindent \textbf{Highest clearance} --- chooses the LR view that has the best clearance score, as in \cite{molini2019deepsum}.

\noindent \textbf{Median} --- takes the median of the $9$ clearest LR observations as the reference, as in \cite{deudon2020highres}.

\noindent \textbf{Heuristic} --- In the test set, the ground truth HR are not available so we use a heuristic to predict the reference images. By minimizing this objective function:
\begin{equation}
    \begin{aligned}
    i_{\text{heur}} &= \text{argmin}_i \, \Big\{\|\text{Mask}^\text{LR}_i - \text{Downscale}(\text{Mask}^\text{HR}) \|_1  \\
    &+ \alpha \left|\text{median}(\text{LR}_i) - \text{median}(\text{LRset})\right| \\
    &+ \beta \,\text{clearance}(\text{LR}_i) \Big \},
    \end{aligned}
\end{equation}
where Mask designates the status map of an image, LRset is the set of input LR images, clearance is the sum of all clear pixels of a LR,
we manage to guess the true references in more than $50\%$ of scenes in the training set.
We set $\alpha = 0.1, \beta = 0.3$ in our experiments. 
\vspace*{-1mm}
\begin{table}
\caption{Average cPSNR (dB) over the validation dataset for DeepSUM and HighRes-net. The original performance is highlighted in orange and the best performances are highlighted in blue}
\begin{center}
\resizebox{\columnwidth}{!}{%
\begin{tabular}{ll|cccc}
\toprule
\multirow{2}{*}{\textbf{Methods}} & \textbf{Training} &\multicolumn{4}{c}{\textbf{Evaluation ref.}} \\
& \textbf{ref.} & \textit{Simil.}& \textit{Clearance}& \textit{Median} & \textit{Heuristic} \\
\midrule
DeepSUM & \textit{Clearance} & \textcolor{blue}{$\mathbf{47.99}$} & \textcolor{orange}{$\mathbf{47.75}$} & $47.62$ & $47.87$\\
HighRes-net & \textit{Median} & \textcolor{blue}{$\mathbf{47.77}$} & $47.26$ & \textcolor{orange}{$\mathbf{47.48}$} & $47.57$\\
\bottomrule
\end{tabular}
\label{tab1}
}
\end{center}
\end{table}

\begin{table}
\caption{Average cPSNR (dB) over the validation dataset for DeepSUM-SI, DeepSUM-ref and HighRes-net-ref. For each methods, the best performance is highlighted in blue.}
\begin{center}
\resizebox{\columnwidth}{!}{%
\begin{tabular}{ll|cccc}
\toprule
\multirow{2}{*}{\textbf{Methods}} & \textbf{Training} &\multicolumn{4}{c}{\textbf{Evaluation ref.}} \\
& \textbf{ref.} & \textit{Simil.}& \textit{Clearance}& \textit{Median} & \textit{Heuristic} \\
\midrule
DeepSUM-ref & \textit{Similarity} &\textcolor{blue}{$\mathbf{50.24}$} & $46.38$ & $46.69$ & $\mathbf{49.10}$ \\
HighRes-net-ref &\textit{Similarity} & \textcolor{blue}{$\mathbf{50.49}$} & $46.35$ & $46.47$ & $\mathbf{49.29}$ \\
DeepSUM-SI & \textit{Similarity} & \textcolor{blue}{$\mathbf{49.05}$} & $45.57$ & $45.85$ & $\mathbf{47.96}$ \\
\bottomrule
\end{tabular}
\label{tab2}
}
\end{center}\vspace{-1em}
\end{table}

\subsection{Discussion}

Inspecting the results (Table~\ref{tab1}), we observe that the two top competitors of the PROBA-V challenge are affected by the type of reference images. 
Without retraining, using the true references or even the ``heuristic references'' systematically improves the results. In this setting, DeepSUM is better than HighRes-net.

Being trained with the true references (Table~\ref{tab2}), DeepSUM-ref and HighRes-net-ref are superior to the original DeepSUM and HighRes-net by a very large margin ($2.49$ and $3.01$ dB). With the ``heuristic references'', they can still 
surpass the original methods by $1.35$ and $1.81$ dB respectively. 
We admit that by using $\text{Mask}^\text{HR}$ this method does not follow the rules of the contest. However, as a proof of concept, we submitted the results of HighRes-net-ref with the ``heuristic references'' on the official post-mortem PROBA-V challenge\footnote{https://kelvins.esa.int/PROBA-v-super-resolution-postmortem}.  At this point in time, the resulting method is ranked
the second place in the leaderboard and surpassing significantly
the performances of the original DeepSUM and HighRes-net. 
Although this heuristic is based on the mask of the HR, it shows the impact that the choice of the reference image can have on the results.

Furthermore, observe that in this situation where the true references are provided, HighRes-net-ref is better than DeepSUM-ref. We can conclude that the design of the challenge strongly affects its outcome.

On the other hand, the SISR algorithm DeepSUM-SI achieves much better results than the MISR algorithm DeepSUM. This is due to the temporal variability between LR observations. In some sense, networks trained without the knowledge of the reference image have to deal with two different tasks: guessing the reference and super-resolving that specific image using the complementary information from other images in the set. Of course the guess is random (at least among the LR images with high clearance), thus the network will predict some sort of average SR image. Adding the information about the reference helps the networks to focus on the super-resolution problem. 
\setlength\belowcaptionskip{-1ex}
\begin{figure} 
    \centering
    \includegraphics[width = 0.97\linewidth]{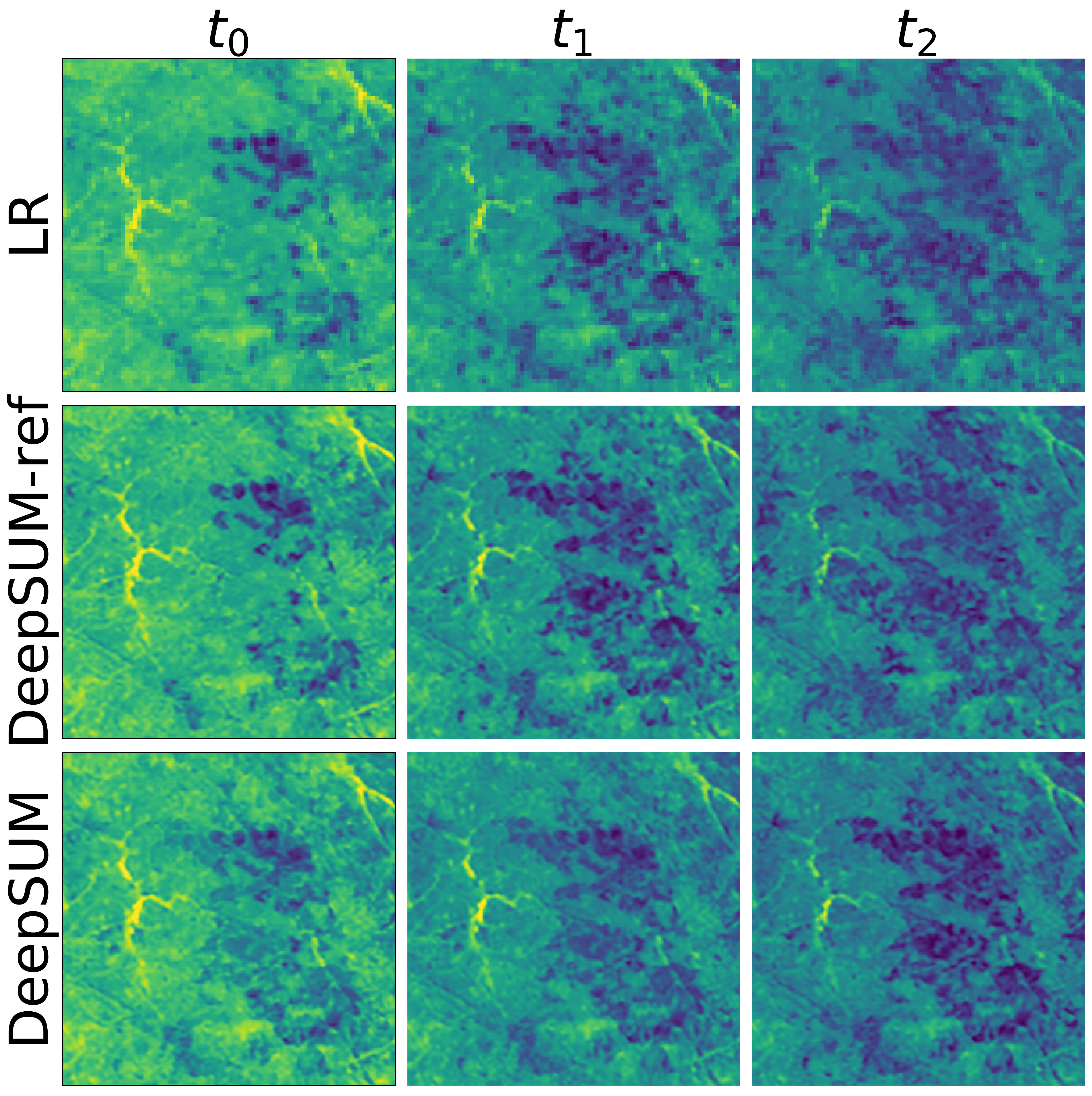}
    \caption{Examples of reconstruction by DeepSUM-ref and DeepSUM with different references (in false color). The first line corresponds to crops of three different LR images in a set. The second line and the third line show the reconstruction by DeepSUM-ref and DeepSUM respectively when using each of these three LR as the reference image.}
    \label{fig:2}
\end{figure}
To evaluate the impact of the reference on the result of DeepSUM and DeepSUM-ref, we select three LR images taken in different days as the reference (see Fig.~\ref{fig:2}). In each case, DeepSUM-ref faithfully recovers fine details in the SR image. On the other hand, the vegetation covers on the outputs of DeepSUM are inconsistent with that of the references. The reconstruction of DeepSUM is less likely to correlate with the reference. Consequently, DeepSUM-ref is more appropriate to practical use of super-resolution since we usually want to super-resolve a specific image in a time series.

%% file: contents/conclusion.tex
\section{Conclusion}

In this work, we have demonstrated that the PROBA-V challenge, by not providing the true LR reference is evaluating not only the MISR performance of the methods, but also the way in which the LR reference images are chosen. The later aspect is irrelevant in the many practical use cases where the reference image is dictated by the application.
To address this use case, we proposed PROBA-V-REF a variant of the dataset with the true reference images in the training and validation splits. These were obtained by comparing the LR images and a downscaled version of the ground truth HR. 
We believe that, by using the provided true LR images, future methods will be able to use this unique real dataset to focus on the core problem of MISR: making the most out of the complementary information in the LR images.